\documentclass[runningheads]{llncs}

\usepackage{eccv}

\usepackage{eccvabbrv}

\usepackage{graphicx}
\usepackage{booktabs}

\usepackage{color}
\usepackage{bm}
\usepackage{multirow}
\usepackage{flushend}
\usepackage{caption}

\usepackage[accsupp]{axessibility}  

\usepackage[pagebackref,breaklinks,colorlinks,citecolor=eccvblue]{hyperref}
\usepackage{hyperref}

\usepackage{orcidlink}

\begin{document}

\title{Match-Stereo-Videos: Bidirectional Alignment for Consistent Dynamic Stereo Matching} 

\titlerunning{BiDAStereo}

\author{Junpeng Jing  \and
Ye Mao \and
Krystian Mikolajczyk }

\authorrunning{J. Jing et al.}

\institute{Imperial College London \\
\url{https://tomtomtommi.github.io/BiDAStereo/} \\
\email{\{j.jing23; ye.mao21; k.mikolajczyk\}@imperial.ac.uk}}

\maketitle

{
\renewcommand\twocolumn[1][]{#1}
\thispagestyle{empty}
    \begin{center}
        \includegraphics[width=1.0\linewidth,page=1]{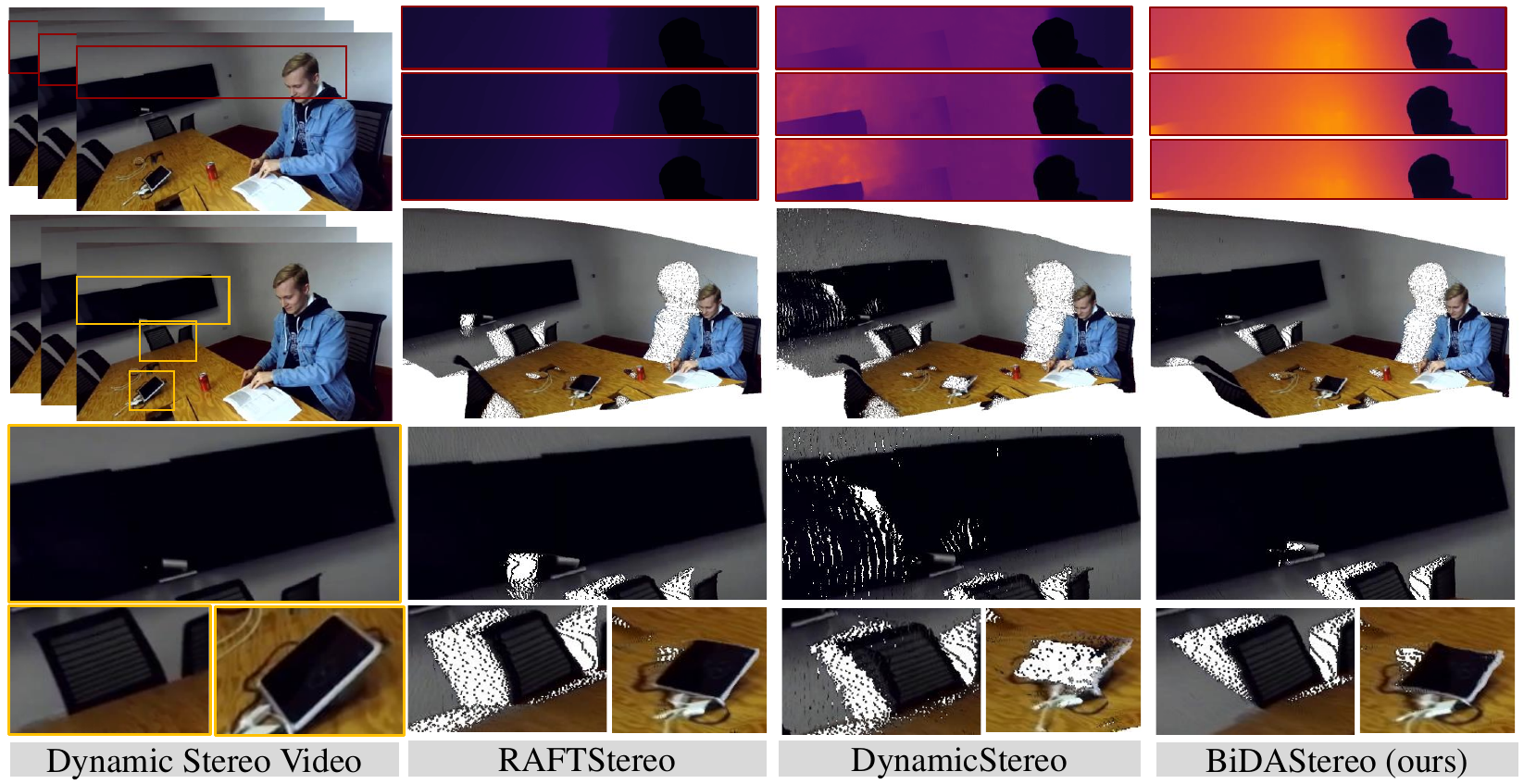}
        \captionof{figure}{Dynamic stereo video. First row: depth maps of the same region in three different frames. Second row: depth maps converted to globally aligned point clouds and rendered with a camera displaced by 15 degree angles. Our method gives consistent and accurate disparities without flickering.}
        \label{fig:figure1}
    \end{center}
}

\begin{abstract}
  Dynamic stereo matching is the task of estimating consistent disparities from stereo videos with dynamic objects.
  Recent learning-based methods prioritize optimal performance on a single stereo pair, resulting in temporal inconsistencies.
  Existing video methods apply per-frame matching and window-based cost aggregation across the time dimension, leading to low-frequency oscillations at the scale of the window size. Towards this challenge, we develop a bidirectional alignment mechanism for adjacent frames as a fundamental operation. We further propose a novel framework, BiDAStereo, that achieves consistent dynamic stereo matching. Unlike the existing methods, we model this task as local matching and global aggregation. Locally, we consider correlation in a triple-frame manner to pool information from adjacent frames and improve the temporal consistency. Globally, to exploit the entire sequence's consistency and extract dynamic scene cues for aggregation, we develop a motion-propagation recurrent unit. Extensive experiments demonstrate the performance of our method, showcasing improvements in prediction quality and achieving state-of-the-art results on various commonly used benchmarks.
\end{abstract}

\section{Introduction}
Stereo matching is a fundamental computer vision task \cite{scharstein2002taxonomy} of estimating the disparity between two rectified stereo images. This task holds significance across diverse applications, including 3D reconstruction \cite{geiger2011stereoscan}, robot navigation \cite{desouza2002vision}, and augmented reality (AR) \cite{azuma1997survey}. It streamlines 3D scene reconstruction, enabling seamless integration into virtual or mixed reality experiences, as well as facilitating mixed reality traversal. As consumer devices such as AR glasses and smartphones equipped with multiple cameras become increasingly prevalent, there is a growing demand for advanced video stereo matching capabilities. 

Deep learning based stereo matching methods have made significant strides in terms of accuracy \cite{lipson2021raft,li2022practical,xu2023iterative}, efficiency \cite{song2021adastereo,xu2022acvnet}, and robustness \cite{shen2021cfnet,Jing_2023_ICCV}. However, achieving consistent disparity estimations from stereo video sequences remains a challenge. Directly applying these methods to video often results in severely flickering disparity maps, since the processing is done on a per-frame basis without considering cross-frame information. As shown in Fig.~\ref{fig:figure1}, RAFTStereo \cite{lipson2021raft}, which excels in performance on image-based stereo benchmarks \cite{eth3d,middlebury,kitti}, struggles to generate consistent disparities on the real-world stereo video. This challenge is amplified in dynamic scenes where objects move and deform. In such scenarios, multi-view constraints \cite{hartley2003multiple} are not applicable and disparities are not translationally invariant. Even if correspondences are established, simply fusing independent disparity maps of corresponding points is ineffective \cite{li2023temporally}.

To this end, some recent approaches proposed to leverage cross-frame information. Li \etal \cite{li2023temporally} first proposed a general pipeline CODD to break down this problem into sub-modules for processing. It consists of a matching network for per-frame disparity estimation, a motion network for SE3 transformation prediction, and another network for temporal information fusion. This method is limited as it only considers temporal information by using one past frame. To expand the temporal receptive field, DynamicStereo \cite{karaev2023dynamicstereo} is designed with self and cross attention mechanisms to extract and pool information over a range of frames based on a transformer architecture. Although it achieves better performance than per-frame methods, it applies per-frame matching and sliding window-based aggregation (Fig.~\ref{fig:figure2} (left)). This mechanism lacks temporal correlation consistency and global sequence information, leading to low-frequency oscillations at the scale of the temporal window size, as shown in Fig.~\ref{fig:figure1}. Moreover, since the position of matching point pairs in stereo videos is changing across the time dimension, directly applying temporal attention in aggregation for different time steps' cost volumes without alignment is sub-optimal. Thus, the key goal for our research is to design a framework capable of effectively leveraging temporal information in the correlation and aggregation process.

\begin{figure}[t]
  \centering
  \includegraphics[width=1\textwidth]{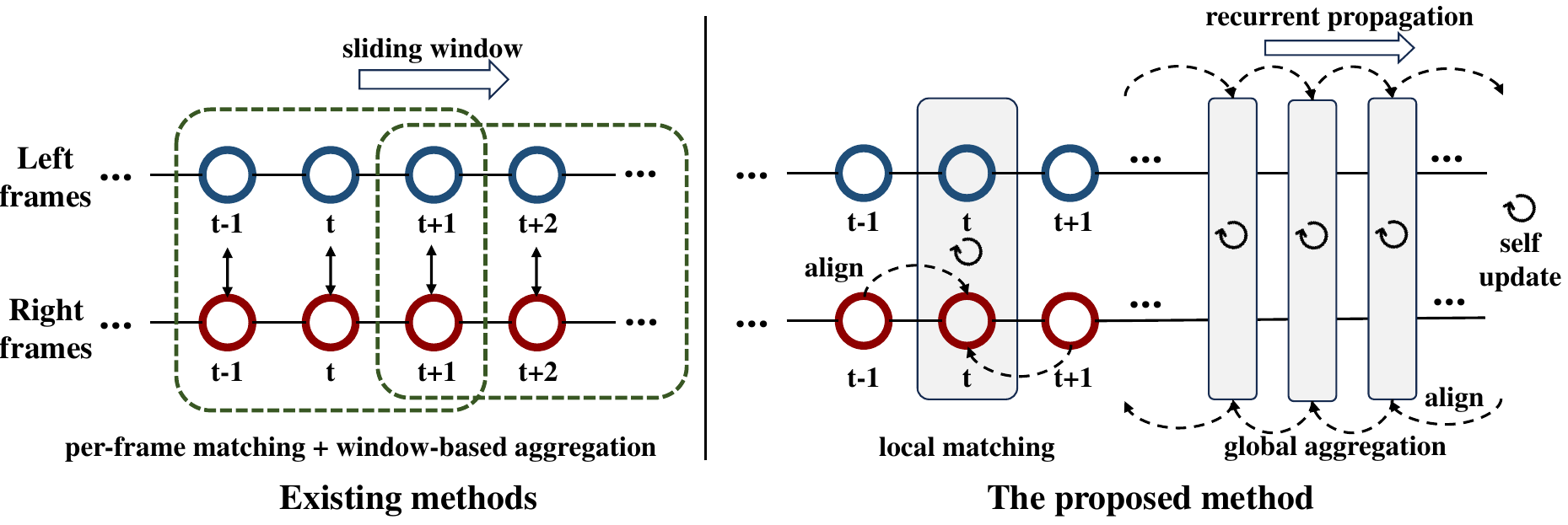}
  \caption{Illustration of the difference between existing methods (left) and our proposed method (right). Existing methods separate video sequences into fixed segments for processing, adopt a per-frame matching operation to build the cost volumes and apply the sliding window for aggregation, thus limiting the information propagation to a fixed time length. Our method adopts bidirectional alignment for local matching, where the cost volumes are built within the neighboring frames. A self-update mechanism is proposed to update the current state via bidirectional alignment and propagate global consistency across the whole sequence. Details of the self-update can be seen in Sec.~\ref{Motion-propagation based Recurrent Unit}.
}
  \label{fig:figure2}
\end{figure}

To tackle these challenges, we highlight the significance of frame alignment in video stereo matching, and develop a novel framework based on Bi-Directional Alignment, namely BiDAStereo, to achieve consistent dynamic stereo matching. As shown in Fig.~\ref{fig:figure2} (right), we apply the bidirectional alignment for two purposes. Firstly, points occluded in one camera at a given time step may become visible from both cameras in adjacent time steps. To leverage temporal information from neighboring frames (local matching), we align the frames towards the center frame and build the cost volumes via a triple-frame correlation layer. Secondly, to exploit information from the entire sequence and extract dynamic cues (global aggregation), we develop a Motion-propagation Recurrent Unit (MRU). Within the MRU updates, bidirectional motion features from neighboring frames are aligned and fused towards the center frame to update the center frame. This approach allows for the recurrent propagation of global consistency, expanding the temporal receptive field and enabling the model to exploit a broader range of temporal information. This provides a significant advantage, particularly in dynamic scenes, as ambiguity and insufficient information in per-frame estimation are mitigated when multiple frame information is effectively utilized, as illustrated in Fig.~\ref{fig:figure1}.

The main contributions of this paper are as follows:
\begin{itemize}
\item
The bidirectional alignment mechanism is developed as an effective operation for enforcing temporal consistency in dynamic stereo vision.
\item
A triple-frame correlation layer is proposed to align adjacent frames and build cost volumes, extracting local temporal receptive field cues. 
\item 
A novel motion-propagation recurrent unit is proposed to exploit the global temporal information in dynamic scenes.
\item
The proposed method achieves SOTA performance on dynamic stereo matching results among a variety of benchmarks.
\end{itemize}

\section{Related Works}
In this section, we first review per-frame stereo matching and then discuss recent methods for stereo video.
\subsection{Per-frame Stereo Matching} 
Stereo matching remains a classic and intricate problem extensively researched in computer vision in recent years. Most of the traditional approaches can be categorized into local and global. Local methods divide images into patches, calculate matching costs between these patches, and subsequently perform local aggregation \cite{birchfield1999depth, hirschmuller2002real, van2002hierarchical}. In contrast, global methods formulate the task as an energy optimization problem, employing an explicit cost function  \cite{sun2003stereo, klaus2006segment, yang2008stereo, boykov2001fast}, which is optimized by belief propagation or graph cut algorithms.

Zbontar and LeCun \cite{zbontar2015computing} were pioneers in leveraging convolutions for matching cost computation. Mayer \etal \cite{mayer2016large} further propelled this field forward by introducing the first end-to-end network. Subsequently, research on end-to-end networks has branched into two main directions. One direction focuses on utilizing 2D convolutions, where innovative mechanisms and modules have been proposed, such as cascaded connection \cite{pang2017cascade}, group-wise correlation \cite{guo2019group}, adaptive aggregation \cite{xu2020aanet}, and hierarchical connection \cite{tankovich2021hitnet}. Alternatively, another direction involves constructing 4D cost volumes and leveraging 3D convolutions for aggregation. Various techniques like 3D hourglass aggregation module \cite{kendall2017end}, spatial pyramid connection \cite{chang2018pyramid}, semi-global guided aggregation \cite{zhang2019ga}, and high-resolution targeted multi-scale layer \cite{yang2019hierarchical} have been introduced. In addition to accuracy, there are also some methods focusing on other properties important in real-world scenarios, such as efficiency \cite{song2021adastereo,xu2022acvnet}, robustness \cite{shen2021cfnet,Jing_2023_ICCV}, and domain generalization  \cite{pang2018zoom,rao2023masked,chang2023domain}.

Most recently, iterative mechanism has showcased its effectiveness \cite{teed2020raft, lipson2021raft}. Building on this, Li \etal \cite{li2022practical} proposed adaptive correlation and incorporated attention mechanism, yielding performance improvements. Xu \etal \cite{xu2023iterative} combined geometry encoding volume with the backbone of \cite{lipson2021raft} and used the results from 3D aggregation network as the initial input to the iterative refinement. Despite these notable strides, these methods primarily address per-frame stereo matching, neglecting temporal information in video sequences. Direct application of these methods to video frames often results in poor temporal consistency, notably manifested as severe flickering in the output disparities.

\subsection{Video Stereo Matching}
Matching stereo videos, particularly with dynamic scenes, is a crucial yet relatively unexplored area. Conventional approaches, exemplified by Patchmatch Stereo \cite{bleyer2011patchmatch}, operate under the assumption of shared planes among neighboring frames, employing 3D spatial windows and temporal propagation to estimate disparities.  Zhong \etal \cite{zhong2018open} introduced an LSTM-based approach for stereo videos, which focused on unsupervised learning for limited real-world annotations, neglecting the temporal information.

Towards dynamic scenes, Li \etal \cite{li2023temporally} first proposed CODD emphasizing temporal consistency. Expanding upon a per-frame stereo network, they introduced separate motion and fusion networks to align and aggregate current and past estimated disparities. Zhang \etal \cite{zhang2023temporalstereo} developed a coarse-to-fine network that leverages past context, enhancing predictions in challenging scenarios such as occlusions and reflective regions. Cheng \etal \cite{cheng2024stereo} contributed to the field by generating a synthetic dataset in an indoor XR scenario and designing a framework to reduce computational costs through temporal cost aggregation. However, these methods are limited in their ability to propagate temporal information solely within the past neighboring frame, thereby constraining the temporal receptive field. Moreover, they heavily rely on auxiliary camera motions or pre-given scene geometries for fusing temporal information, restricting the choice of training datasets. Karaev \etal \cite{karaev2023dynamicstereo} enlarged the receptive field by introducing a transformer-based architecture featuring time, stereo, and temporal attention mechanisms. Despite achieving improved results through sliding window processing, temporal information remains confined within a fixed pre-set range, falling short of encompassing the entire sequence. Additionally, lacking alignment operations, it proves suboptimal for cost aggregation.

\section{Methods}

\begin{figure}[t]
  \centering
  \includegraphics[width=1\textwidth]{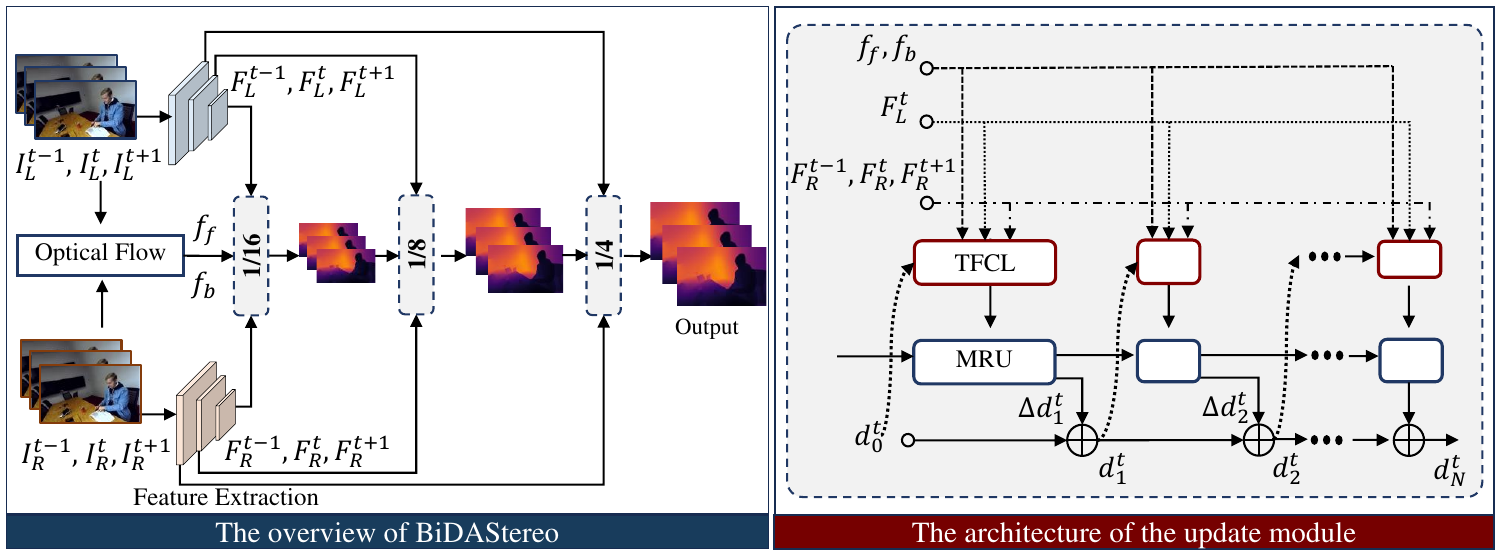}
  \caption{\textbf{Left:} The overall pipeline of the proposed method. Given a pair of stereo sequences, bidirectional optical flows are estimated and feature maps are extracted at three scales. In each scale, the predicted disparities are refined iteratively in the update module, and the final output of the former stage is fed to the next one as an initialization. The same update module is reused in each stage. \textbf{Right:} The architecture of the update module. For each iteration, the Triple-Frame Correlation Layer (TFCL) is used to compute cost volumes from triple-frame feature maps. The motion-propagation Recurrent Unit (MRU) is used for global cost aggregation and disparity estimations.}
  \label{fig:framework}
\end{figure}

Given a pair of rectified stereo sequences $\{\mathbf{I}^{t}_{L}, \mathbf{I}^{t}_{R}\}_{t\in (1,T)} \in \mathbb{R}^{H\times W\times 3} $, the task of dynamic stereo matching is to estimate a sequence of disparity maps $\{\mathbf{d}^{t}\}_{t\in (1,T)} \in \mathbb{R}^{H\times W} $ aligned with the left one, where $T$ is the number of frames. The challenge lies in devising a model that is capable of efficiently propagating consistency across the whole sequence. Existing methods often apply per-frame matching and window-based aggregation across frames $T$, neglecting the fundamental alignment operation between adjacent frames. Since the position of matching points in stereo images is moving across time, relying solely on temporal attention without alignment is sub-optimal.
Towards this limitation, we propose BiDAStereo based on bidirectional alignment for dynamic stereo matching. As illustrated in Fig.~\ref{fig:framework}, the framework consists of three modules: a feature extraction module, an optical flow module, and an update module for disparity estimations.

To simplify the process, we exemplify the input with three frames (the center and its neighboring frames) but the same approach can be applied to the cases with more than three frames. In the left part of Fig.~\ref{fig:framework}, the input stereo sequences are processed by two shared-weight convolutional feature extraction modules. Multi-scale feature maps $\left \{\mathbf{F}_{L}^t ,\mathbf{F}_{R}^t \right \}_s \in R^{sH\times sW\times C}$ are extracted, where $s \in \left \{1/16, 1/8, 1/4\right \}$ represents down-sampled scales, and $C$ is the channel number. Bidirectional optical flow $\{\mathbf{f}_f, \mathbf{f}_b \} \in \mathbb{R}^{H\times W\times 2} $ are simultaneously estimated and downsampled to the corresponding resolution of the feature maps. Subsequently, the extracted features and the optical flow maps traverse three cascaded stages of the update module, which consists of a Triple-Frame Correlation Layer (TFCL) and a Motion-propagation Recurrent Unit (MRU). In the TFCL, cost volumes are built using the adjacent aligned features and input into the MRU, iteratively refining disparity predictions. Initialized from a blank disparity map, the output disparities from the previous update stage are fed into the next stage. The same update module is used in each stage to reduce overall parameters. Finally, the predicted disparities at the last stage are re-scaled to the original resolution using convex up-sampling \cite{teed2020raft}.

\begin{figure}[t]
  \centering
  \includegraphics[width=1\textwidth]{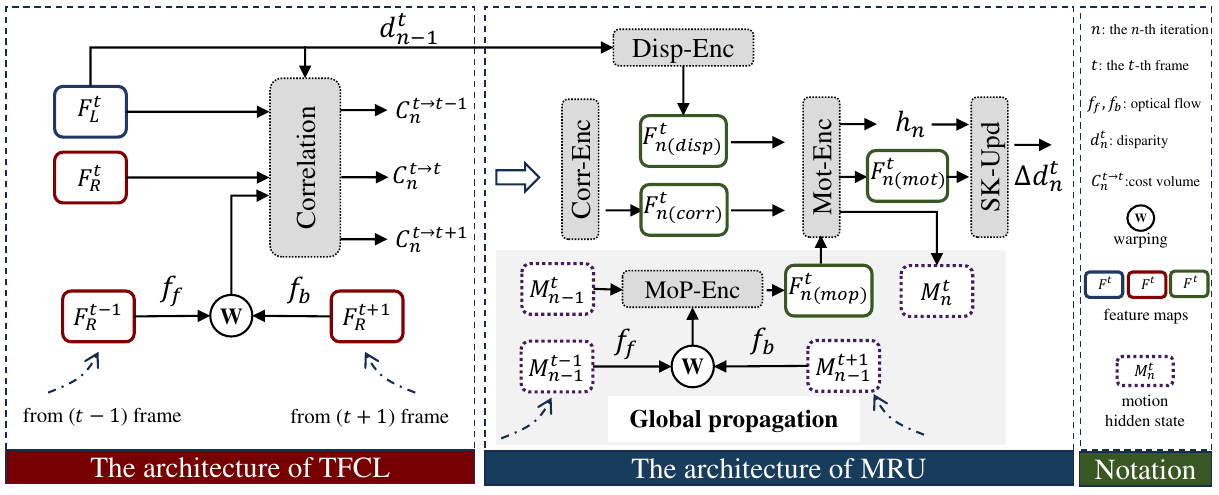}
  \caption{The architecture of TFCL and MRU. For TFCL, bidirectional alignment is conducted from the adjacent right frames to the center frame. Cost volumes are built among the left frame and the aligned right frame. For MRU, convolutional encoders are adopted for correlations, disparities, and motion features. A motion hidden state feature is introduced for each frame as an auxiliary context in global propagation. In each iteration, adjacent motion hidden state features are aligned towards the center one, updating the center one and propagating wider temporal information.}
  \label{fig:module}
\end{figure}

\subsection{Triple-Frame Correlation Layer} 

\noindent \textbf{Bidirectional alignment.} Building cost volumes among multiple frames can enrich the perception information of scenes since occluded points in a given frame could be visible in the adjacent frames. Directly establishing multi-frame correlation is a naive approach and sub-optimal for cost aggregation, since the unaligned features have different position thus introduce noise. Therefore, alignment is indispensable to ensure that corresponding matching points in different frames are positioned identically. As shown in the left part of Fig.~\ref{fig:module}, given the bidirectional optical flow maps in corresponding resolution $\{\mathbf{f}_f, \mathbf{f}_b\}$, the right features $\left \{\mathbf{F}_{R}^{t-1}, \mathbf{F}_{R}^{t+1} \right \}$ are first warped towards the center frame $\left \{\mathbf{F}_{R}^{t}\right \}$ via bilinear warping $\mathcal{W}$ to obtain aligned features $\left \{\hat{\mathbf{F}}_{R}^{t-1}, \hat{\mathbf{F}}_{R}^{t+1} \right \}$, formulated as follows,
\begin{equation}
\hat{\mathbf{F}}_{R}^{t-1} = \mathcal{W}(\mathbf{F}_{R}^{t-1}, \mathbf{f}_f),  \quad  \quad \hat{\mathbf{F}}_{R}^{t+1} = \mathcal{W}(\mathbf{F}_{R}^{t-1}, \mathbf{f}_b).
\end{equation}

\noindent \textbf{Correlation.} Following bidirectional alignment, cost volumes are constructed employing the local correlation mechanism \cite{li2022practical}. We extend this mechanism to a triple-frame version. In particular, for the $n$-th iteration, the intermediate disparity map $\mathbf{d}^t_{n-1}$ from the previous iteration is employed to warp right features towards the left feature map: 
\begin{equation}
[\tilde{\mathbf{F}}_{R}^{t-1}, \tilde{\mathbf{F}}_{R}^{t}, \tilde{\mathbf{F}}_{R}^{t+1}] = \mathcal{W}([\hat{\mathbf{F}}_{R}^{t-1}, {\mathbf{F}}_{R}^{t}, \hat{\mathbf{F}}_{R}^{t+1}], \mathbf{d}^t_{n-1}).
\end{equation} 
Then, the cost volumes $\mathbf{C}_{n}^{t \to t'}(\bm{p})$ between the left feature  map $\mathbf{F}_{L}^{t}$ and the warped right features $\tilde{\mathbf{F}}_{R}^{t'}$ at position $\mathbf{p}$ can be formulated as follows,
\begin{equation}
\mathbf{C}_{n}^{t \to t'}(\bm{p}) = \underset{\mathbf{r} \in \mathbf{R}}{\text{Concat}} \{ \langle \mathbf{F}_{L}^{t}(\mathbf{p}) \cdot  \tilde{\mathbf{F}}_{R}^{t'}(\mathbf{p} + \mathbf{r}) \rangle \},
\end{equation} 
where $t' \in \left \{t-1, t, t+1\right \}$. $\mathbf{R}$ is the search range of the current pixel, where $(\pm 4,0)$ and $(\pm 1,\pm 1)$ are alternatively adopted for horizontal and vertical directions in practical implementation. $\left \langle \cdot \right \rangle $ represents channel-wise product operation. 

It is worth noting that only a single left frame is utilized in the correlation layer. In multiple-frame matching, constructing cost volumes across different source (right) frames increases potential matching options as auxiliary information. Conversely, considering multiple reference (left) frames merely results in ambiguity, lacking a clear physical location of features and potentially introducing noise that can impact the matching performance. This is further corroborated by the ablation study in Section \ref{Ablation Studies}.

\subsection{Motion-propagation Recurrent Unit} \label{Motion-propagation based Recurrent Unit}
With TFCL, the model can utilize temporal information from neighboring two frames. However, the extent of information propagation is still confined to a local range. To facilitate the propagation of global consistency, it's essential to establish a state variable representing each frame and fuse it across adjacent frames. While feature maps from images are commonly used as state variables in tasks like video super-resolution and denoising \cite{liu2022video}, they are not well-suited for stereo matching, as fusion with other frames could significantly impact matching accuracy. Instead, motion features, which provide context, are more appropriate for global propagation and dynamic scenes. As illustrated in Fig.~\ref{fig:module}, we propose a motion-propagation mechanism designed to exploit contextual information from the entire sequence. During the update iteration, a motion hidden state $\mathbf{M}$ is introduced to cache the state for each frame. This involves bidirectional alignment of neighboring states to iteratively update itself. As a result, information in each frame undergoes recurrent propagation throughout the entire sequence. Detailed elaboration on each module will be provided in subsequent subsections.

\noindent \textbf{Motion propagation.} A motion hidden state feature $\mathbf{M}^{t} \in R^{sH\times sW\times C_0}$ is introduced for the $t$-th frame in the MRU, where $C_0$ is the channel number. For the first iteration at the first stage, $\mathbf{M}^{t}_{0}$ is randomly initialized and undergoes learnable updates. For the $n$-th iteration, the motion hidden state features $\mathbf{M}_{n-1}^{t-1}$ and $\mathbf{M}_{n-1}^{t+1}$ from adjacent frames are aligned towards the center one $\mathbf{M}_{n-1}^{t}$ via bidirectional optical flow maps $\{\mathbf{f}_f, \mathbf{f}_b\}$, formulated as follows, 
\begin{equation}
\hat{\mathbf{M}}_{n-1}^{t-1} = \mathcal{W}(\mathbf{M}_{n-1}^{t-1}, \mathbf{f}_f),  \quad  \quad \hat{\mathbf{M}}_{n-1}^{t+1} = \mathcal{W}(\mathbf{M}_{n-1}^{t+1}, \mathbf{f}_b).
\end{equation}
Then, the aligned features $\{\hat{\mathbf{M}}_{n-1}^{t-1}, \hat{\mathbf{M}}_{n-1}^{t+1}\}$  and the center one $\mathbf{M}_{n-1}^{t}$ are fused by a motion-propagation encoder to obtain the motion propagation feature $\mathbf{F}_{n (mop)}^{t}$:
\begin{equation}
\mathbf{F}_{n (mop)}^{t} = \text{MoP-Enc}(\hat{\mathbf{M}}_{n-1}^{t-1}, \mathbf{M}_{n-1}^{t}, \hat{\mathbf{M}}_{n-1}^{t+1}).
\end{equation}
Across various update scales, the motion hidden state is not re-initialized, and the final output hidden state from the previous level is up-sampled as the initial input for the next level.

\noindent \textbf{Encoder blocks.} After building the cost volumes $\mathbf{C}_{n}^{t \to t'}(\bm{p})$, the correlation feature $\mathbf{F}_{n (corr)}^{t}$ is obtained using a convolutional correlation encoder (Corr-Enc). The disparity encoder (Disp-Enc) is applied to acquire $\mathbf{F}_{n (disp)}^{t}$:
\begin{equation}
\begin{aligned}
\mathbf{F}_{n (corr)}^{t} &= \underset{t' \in \left \{t-1, t, t+1\right \}}{\text{Corr-Enc}}(\mathbf{C}_{n}^{t \to t'}(\bm{p})),\\
\mathbf{F}_{n (disp)}^{t} &= \text{Disp-Enc}(\mathbf{d}_{n-1}^{t}).
\end{aligned}
\end{equation} 
Subsequently, these two features are concatenated with the motion propagation feature $\mathbf{F}_{n (mop)}^{t}$ and processed by the motion encoder (Mot-Enc) to generate the motion feature $\mathbf{F}_{n (mot)}^{t}$, hidden state $\mathbf{h}_n$ \cite{lipson2021raft}, and the updated motion hidden state feature $\hat{\mathbf{M}}_{n}^{t}$:
\begin{equation}
\begin{aligned}
\mathbf{F}_{n (mot)}^{t}, \mathbf{h}_n, \hat{\mathbf{M}}_{n}^{t} &= \text{Mot-Enc}(\mathbf{F}_{n (corr)}^{t}, \mathbf{F}_{n (disp)}^{t}, \mathbf{F}_{n (mop)}^{t}).
\end{aligned}
\end{equation} 

\noindent \textbf{Super kernel updater.} The hidden state $\mathbf{h}_n$ and the motion feature $\mathbf{F}_{n (mot)}^{t}$ are passed through a super kernel updater to get the predicted disparities residuals $\Delta \mathbf{d}_n^t$, which are added to the current disparities $\mathbf{d}_{n-1}^t$ and obtained $\mathbf{d}_{n}^t$, iteratively refining the disparity predictions. Motivated by \cite{karaev2023dynamicstereo}, 3D convolutions are used in the updater for video sequences to enhance temporal consistency. Different from the regular updater with kernel size $1 \times 1 \times 5$, we introduce an extra convolution layer with a super kernel size $1 \times 1 \times 15$. The proposed layer is only adopted for the horizontal direction to expand the receptive field along the epipolar line.

\subsection{Discussion of Alignment}

The alignment of different time steps can be regarded as multi-view stereo matching (MVS), which utilizes explicit geometry information. In MVS, the plane sweep algorithm \cite{yao2018mvsnet} is commonly employed to project source images onto fronto-parallel planes of the reference image using homography. However, due to the presence of dynamic objects, the triangulation rule typically applied in MVS becomes inapplicable. Furthermore, MVS methods necessitate camera pose information and a pre-calculated depth range obtained through Structure from Motion (SfM). The estimation of these items would significantly augment the complexity of the overall pipeline and accumulate errors.

An alternative approach is leveraging scene flow \cite{teed2021raft}, which estimates pixel-wise 3D motions and obviates the need for camera pose estimation. Alignment can be achieved through a SE3 transformation, akin to the method proposed in \cite{li2023temporally}. However, estimating scene flow requires depth of source images or Lidar information, thereby leading to a chicken-and-egg problem for our task. Although iterative optimization techniques can be employed to tackle this issue, the final performance is largely contingent upon the accuracy of the scene flow estimation, as demonstrated in \cite{li2023temporally}.

In comparison to the aforementioned explicit methods, optical flow can be viewed as an implicit alignment method devoid of geometry guidance. It establishes correspondences between 3D points in two camera coordinates directly, making it suitable for both dynamic and static objects. In our method, optical flow aligns adjacent features in the TFCL and serves as a verification mechanism for correlation. Consequently, matching points in adjacent frames should occupy the same position post-alignment. Additionally, it facilitates the connection of adjacent motion features, ensuring seamless motion-propagation updates.

\subsection{Loss Function}
In the training process, $T$ frames are used, and for each frame, $N$ disparity predictions are generated after iterations. All disparity predictions are supervised by $l_1$ distance with ground truth disparities in an end-to-end manner. The last disparity prediction is selected as the final output. The total loss is formulated as follows:
\begin{equation}
\mathcal{L} = \sum_{t=1}^{T} \sum_{n=1}^{N} 
\gamma^{N - n} || \mathbf{d}_{\mathrm{gt}}^{t} - \mathbf{d}_{n}^{t} ||,
\end{equation}
where $\gamma$ is set as 0.9 and $\mathbf{d}_{\mathrm{gt}}^{t}$ is the ground truth for the $t$-th frame. Upsampling is used for lower resolution disparities to the resolution of ground truth.

\section{Experiments}
This section presents BiDAStereo's evaluation datasets and implementation details, demonstrates its out-of-domain and in-domain temporal consistency, and performs ablation studies to confirm the effectiveness of its components.

\subsection{Datasets}
For training, SceneFlow \cite{mayer2016large} and Dynamic Replica training set \cite{karaev2023dynamicstereo} are used. For evaluation, following the previous methods, we adopt the commonly used benchmarks, including Sintel clean and final pass \cite{sintel} and the first 150 frames of Dynamic Replica test set \cite{karaev2023dynamicstereo}.

\noindent \textbf{SceneFlow} (SF) is comprised of three subsets: FlyingThings3D, Driving and Monkaa. FlyingThings3D is an abstract dataset featuring moving shapes against colorful backgrounds. It includes 2,250 sequences, each with 10 frames. Driving involves 16 sequences depicting driving scenarios, with each sequence having 300 to 800 frames. Monkaa consists of 48 sequences in cartoon scenarios, with frame counts ranging from 91 to 501.

\noindent  \textbf{Dynamic Replica} (DR) introduced in \cite{karaev2023dynamicstereo}, is notable for its longer sequences and inclusion of non-rigid objects like animals and people. The dataset includes 484 training sequences each with 300 frames, 20 validation sequences each with 300 frames, and 20 test sequences each with 900 frames.

\noindent  \textbf{Sintel} is derived from computer-animated movies and comprises 23 sequences for both clean and final passes. Each sequence contains 20 to 50 frames.

\subsection{Implementation Details} \label{Implementation Details}
We implement BiDAStereo in PyTorch and train on NVIDIA A100 GPUs. RAFT \cite{teed2020raft} is used for the optical flow module. We first pretrain the model with a batch size of $16$ and resolution $256 \times 256$, and further finetune it with a batch size of $8$ and resolution $256 \times 512$. We use AdamW optimizer \cite{adam} with a standard learning rate of 0.0004 and the default setting. We also adopt the one-cycle learning rate schedule \cite{smith2019super}.  The pretraining process is set to $60k$ iterations for the SF version and $80k$ iterations for the SF+DR version. The finetuning process is set to $40k$ iterations for the SF version and $60k$ iterations for the SF+DR version. The whole training process takes about $5$ days. The optical flow module is frozen in pretraining and trainable in finetuning. Following \cite{karaev2023dynamicstereo}, multiple data augmentation techniques including random crop, rescaling, and shifts in saturation are used in training. The length of the sequences is set to $T=5$ during training, $T=20$ for DR, and full sequence length for Sintel evaluation. The number of iterations is set to $10$ in training and $20$ in evaluation. 

\begin{table}[tb] 
\caption{Out-of-domain evaluation on Sintel clean and final pass datasets \cite{sintel}. SF - SceneFlow \cite{mayer2016large}, K - KITTI \cite{menze2015object}, M - Middlebury \cite{scharstein2014high}, DR - Dynamic Replica \cite{karaev2023dynamicstereo}. 7 datasets include SceneFlow \cite{mayer2016large}, Sintel \cite{sintel}, Falling Things \cite{fallingthings}, InStereo2K
\cite{bao2020instereo2k}, Carla \cite{deschaud2021kitti}, AirSim \cite{airsim2017fsr}, and CREStereo dataset \cite{li2022practical}. Lower values are better for all metrics.}
\centering
\setlength{\tabcolsep}{2.pt}
\scriptsize
\begin{tabular}{c|c|cccc|cccc}
\toprule
\multirow{3}{*}{Training Data} & \multirow{3}{*}{Method} & \multicolumn{8}{|c}{Sintel}                                \\
\cmidrule(lr){3-10}
                                &                         & \multicolumn{4}{c}{clean}   & \multicolumn{4}{c}{final}   \\
\cmidrule(lr){3-10}
                                &                         & $\delta_{1px}^{t}$ & $\delta_{3px}^{t}$ & TEPE & $\delta_{3px}$  & $\delta_{1px}^{t}$ & $\delta_{3px}^{t}$ & TEPE & $\delta_{3px}$  \\
\midrule
\multirow{4}{*}{SF}             & CODD \cite{li2023temporally} & 10.78 & 5.65 & 1.44  & 8.68 & 18.56 & 9.79 & 2.32 & 17.5 \\
                                & RAFTStereo \cite{lipson2021raft}              & 9.33  & 4.51 & 0.92 & 6.12  & 13.69 & 7.08 & 2.10 & 10.4 \\
                                & DynamicStereo \cite{karaev2023dynamicstereo}          & \underline{8.41}  & \underline{3.93} & \underline{0.77} & \underline{6.10} & \underline{11.95} & \underline{5.98} & \underline{1.45}  & \underline{8.97}  \\
                                & BiDAStereo (ours)        & \textbf{8.29} & \textbf{3.79} & \textbf{0.73} &  \textbf{5.94} & \textbf{11.65} & \textbf{5.53} & \textbf{1.26} & \textbf{8.78} \\
								
\midrule
\multirow{3}{*}{SF + DR}        & RAFTStereo \cite{lipson2021raft}             & 9.07  & 4.40 & 0.89 & 5.83 & 13.56 & 7.02 & 1.91 & 9.83  \\
                                & DynamicStereo \cite{karaev2023dynamicstereo}           & \underline{8.46}  & \underline{3.93} & \underline{0.76} & \underline{5.77} & \underline{11.93} & \underline{5.92} & \underline{1.42} & \underline{8.68} \\
                                & BiDAStereo (ours)        & \textbf{8.03}  & \textbf{3.76} & \textbf{0.75} & \textbf{5.75} & \textbf{11.04} & \textbf{5.30} & \textbf{1.22} & \textbf{8.52}  \\
\midrule
SF + M + K                      & CODD \cite{li2023temporally}        & 12.16 & 6.23 & 1.33 & 9.11 & 16.16 & 8.64 & 2.01 & 11.90 \\
SF + M                          & RAFTStereo \cite{lipson2021raft}     & 8.79  & 4.13 & 0.85 & 5.86  & 12.40 & \underline{6.23} & \underline{1.63} & \underline{8.47}  \\
7 datasets (\textbf{incl. Sintel})       & CREStereo \cite{li2022practical}      & \textbf{6.36}  & \textbf{3.26} & \textbf{0.67} & \textbf{4.58} & \underline{12.29} & 6.87 & 1.90 & \textbf{8.17}  \\
SF + DR                         & BiDAStereo (ours)        & \underline{8.03}  & \underline{3.76} & \underline{0.75} & \underline{5.75}  & \textbf{11.04} & \textbf{5.30} & \textbf{1.22} & {8.52} \\

\bottomrule
\end{tabular}
\label{tab:sintel-results}

\end{table}

\begin{table}[htb]
\caption{In-domain evaluation on the Dynamic Replica test set  \cite{karaev2023dynamicstereo}. }
\centering
\setlength{\tabcolsep}{8.pt}
\scriptsize
\begin{tabular}{c|c|cccc}
\toprule
\multirow{2}{*}{Training Data} & \multirow{2}{*}{Method}               & \multicolumn{4}{c}{Dynamic Replica (first 150 frames)}  \\
\cmidrule(lr){3-6}
                                &                                       & $\delta_{1px}^{t}$ & $\delta_{3px}^{t}$ & TEPE  & $\delta_{1px}$   \\
\midrule
\multirow{3}{*}{SF + DR}        & RAFTStereo \cite{lipson2021raft}       & 0.84    & 0.27 & 0.082 & \textbf{1.88}     \\
                                & DynamicStereo \cite{karaev2023dynamicstereo}     & \underline{0.68}    & \underline{0.23} & \underline{0.075} & 3.32   \\
                                & BiDAStereo (ours)                           & \textbf{0.61}    & \textbf{0.22}  & \textbf{0.062} & \underline{2.81}   \\
\midrule
SF + M + K                      & CODD \cite{li2023temporally}            & 2.16    & 0.77  & 0.152 & 10.03    \\
SF + M                          & RAFTStereo \cite{lipson2021raft}         & 1.34    & 0.41 & 0.114 & 3.46    \\
7 datasets (incl. Sintel)       & CREStereo \cite{li2022practical}        	& \underline{0.88}	  & \underline{0.29} & \underline{0.088} & \textbf{1.75} 	  \\
SF + DR                         & BiDAStereo (ours)                         & \textbf{0.61}    & \textbf{0.22}  & \textbf{0.062} & \underline{2.81}   \\
\bottomrule
\end{tabular}
\label{tab:dr-results}
\end{table}

\subsection{Temporal Consistency}
To evaluate temporal consistency, we compute the temporal end-point-error (TEPE) \footnote{\scalebox{.9}{$\mathrm{TEPE}(\mathbf{d}, \mathbf{d}_{\mathrm{gt}})=\sqrt{\sum_{t=1}^{T-1}((\mathbf{d}^{t} - \mathbf{d}^{t+1}) - (\mathbf{d}_{\mathrm{gt}}^{t} - \mathbf{d}_{\mathrm{gt}}^{t+1}))^{2}} $}}, which measures the variation of the end-point-error across time dimension. $\delta_{n-px}^{t}$ represents the proportion of pixels with TEPE higher than the $n$ threshold. Lower values on both metrics indicate greater temporal consistency.

\begin{figure}[t]
  \centering
  \includegraphics[width=.99\textwidth]{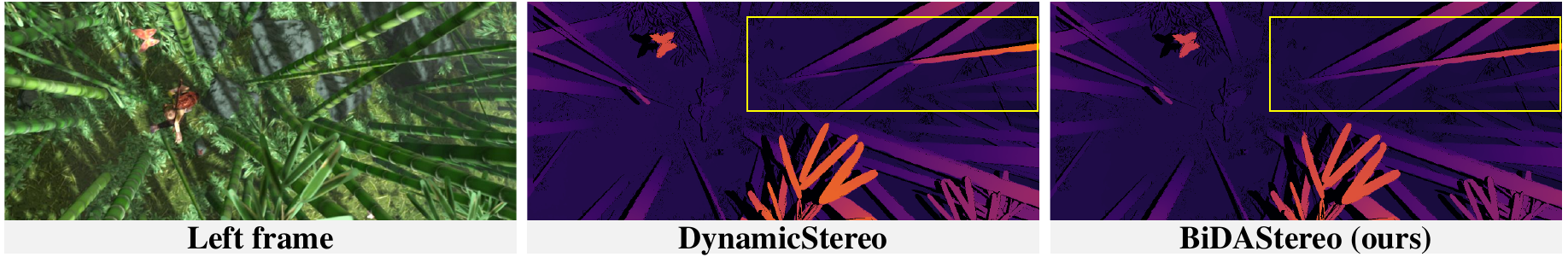}
  \caption{Qualitative comparisons on Sintel final dataset \cite{sintel}.}
  \label{fig:sintel-comparison}
\end{figure}

\begin{figure}[t]
  \centering
  \includegraphics[width=.98\textwidth]{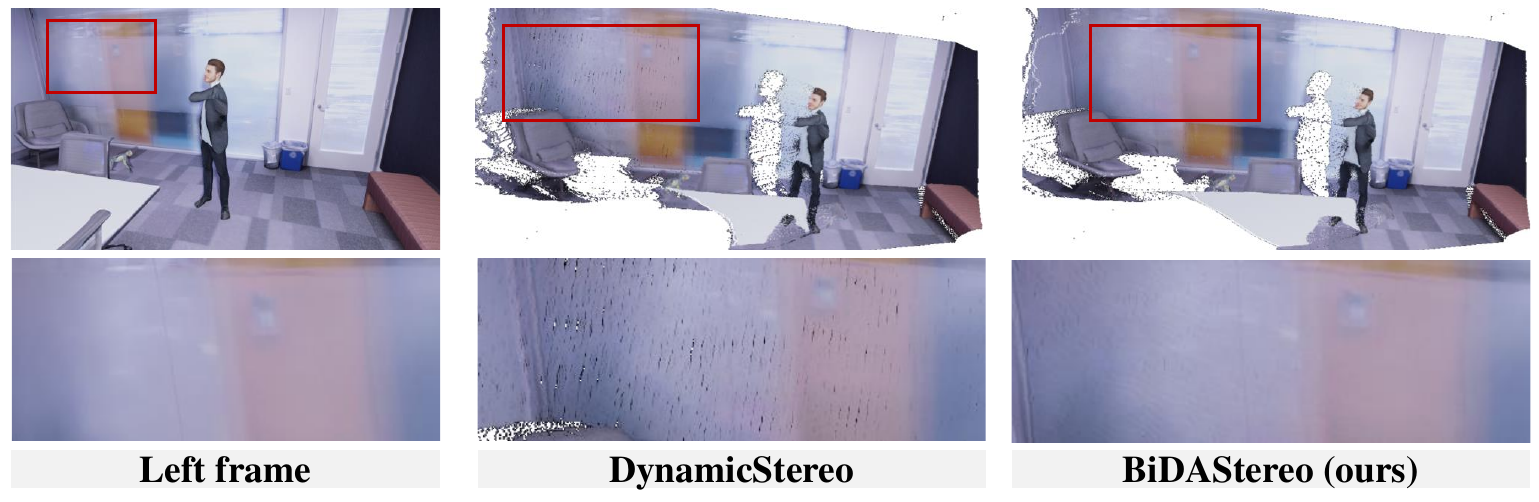}
  \caption{Qualitative comparisons on Dynamic Replica test set \cite{karaev2023dynamicstereo}.}
  \label{fig:dr-comparison}
\end{figure}

\noindent \textbf{Out-of-domain evaluation.} In Tab.~\ref{tab:sintel-results}, when trained on the SceneFlow dataset, our method gives the best results in terms of accuracy and temporal consistency. Specifically, our BiDAStereo outperforms DynamicStereo \cite{karaev2023dynamicstereo} by 5.2\% and 13.1\%; RAFTStereo \cite{lipson2021raft} by 20.7\% and 40.0\% in TEPE on Sintel clean and final pass, respectively. For models trained with SceneFlow and Dynamic Replica, our method achieves the top rank across all metrics, with an improvement of 14.1\% and 36.1\% in TEPE on the Sintel final pass compared to DynamicStereo and RAFTStereo. The incorporation of Dynamic Replica as an additional training set enhances the temporal consistency of all models. Notably, our method exhibits better performance on the final pass, underscoring its enhanced robustness. We further present the results of the methods trained with diversified datasets. Compared to CODD \cite{li2023temporally} and RAFTStereo trained with real-world per-frame datasets Middlebury and KITTI, our method trained solely on synthetic datasets still demonstrates superior performance. Moreover, BiDAStereo exhibits even better temporal consistency than CREStereo \cite{li2022practical} on the final pass, which is a per-frame method trained including the Sintel test set. The qualitative comparative results on Sintel final pass are presented in Fig.~\ref{fig:sintel-comparison}.

\noindent \textbf{In-domain evaluation.} As shown in Tab.~\ref{tab:dr-results}, our method outperforms prior methods on temporal consistency, achieving 0.062 TEPE on the Dynamic Replica test set. Trained on two synthetic datasets, our method shows better temporal performance compared to CREStereo \cite{li2022practical}, which is trained on 7 diversified datasets. This indicates the effectiveness of utilizing temporal information. In Fig.~\ref{fig:dr-comparison}, we visualize the rendered images using the disparity predictions of DynamicStereo \cite{karaev2023dynamicstereo} and our BiDAStereo to show the superior performance of our approach. The red boxes highlight where our method handles ambiguity better especially in areas with weak textures such as glass. In contrast, DynamicStereo produces blurry predictions with visible distortions. The results further support our claim that employing local matching and global aggregation via bidirectional alignment can guide the network to  enforce temporal consistency.

\begin{figure}[t]
  \centering
  \includegraphics[width=1\textwidth]{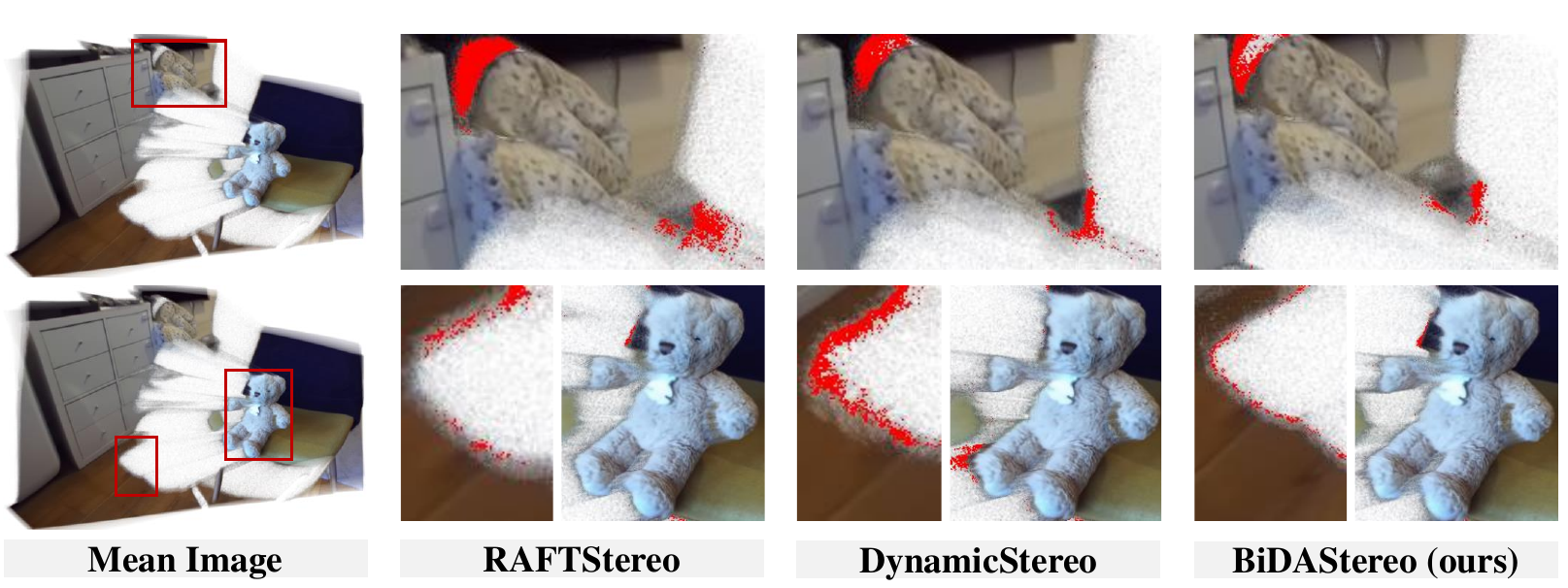}
  \caption{Temporal consistency comparisons. Visualization on a 40-frame reconstructed real-world static stereo video.  All models are trained on DR \& SF. We first predict disparity for each frame and convert it to globally aligned point clouds. They are further rendered with a camera displaced by 15 degree angles. Mean and variance across all images are computed and pixels with variance higher than 40 px2 are shown in red. Our method shows a lower variance, indicating better consistency.}
  \label{fig:real-comparison}
\end{figure}

\noindent \textbf{Real-world static scene.} Although our method is designed for dynamic stereo scenes, following \cite{karaev2023dynamicstereo}, we present a qualitative comparison of temporal consistency on a real-world static scene. In Fig.~\ref{fig:real-comparison}, images are converted to point clouds using depth predictions and rendered with a camera displaced by 15 degree angles. We compute the mean and variance across image reconstructions. We color the pixels of the mean image with variance higher than 40 px2 into red. Smaller red regions illustrate lower variance thus better consistency of our method.

\begin{table}[tb]
\caption{Ablation Study on Sintel datasets \cite{sintel} and Dynamic Replica \cite{karaev2023dynamicstereo}. Settings used in the final model are bold. ($+$) represents combined with all the above bold settings. ($-$) represents the experiment that has been conducted but not adopted to the final model. Lower values indicate better results for all metrics. See Sec.~\ref{Ablation Studies} for details.
}
\centering
\setlength{\tabcolsep}{1.pt}
\scriptsize
\begin{tabular}{l|c|cccc|cccc|c}
\toprule
\multirow{2}{*}{Method} & \multirow{2}{*}{Experiment}     & \multicolumn{4}{c}{Sintel Final}   & \multicolumn{4}{|c|}{Dynamic Replica} & \multirow{2}{*}{Parameters}  \\
\cmidrule(lr){3-10}
                                &                         & $\delta_{1px}^{t}$ & $\delta_{3px}^{t}$  & TEPE & $\delta_{3px}$   & $\delta_{1px}^{t}$ & $\delta_{3px}^{t}$ & TEPE & $\delta_{1px}$ &  \\
\midrule
\multicolumn{2}{l|}{Baseline (2D Conv; $1 \times 5$; Shared)}          & 12.83 & 6.50 & 1.74 & 8.45 &	0.84 & 0.27 & 0.091 & 2.40 & 5.0M          \\
\midrule
\multirow{1}{*}{($+$) Updater} & \textbf{3D Conv} & 11.22 & 5.61 & 1.34 &  8.64 & 0.69 & 0.23 & 0.082 & 3.38  & 5.4M \\
\midrule
\multirow{2}{*}{($+$) Kernel size}  & $1\times 3 \times 3$ & 13.41 & 5.99 & 1.43 & 8.93  & 0.91  & 0.27 & 0.087 & 3.48 & 6.3M \\
                                    & \bm{$1\times 1 \times 15$} & 11.19 & 5.56 & 1.33 & 8.59 & 0.69 & 0.23 & 0.079 & 3.26 & 6.9M \\
\midrule
\multirow{1}{*}{($-$) Weights} & Separated  & 11.17 & 5.61 & 1.34 & 8.57 & 0.69 & 0.23 & 0.079 & 3.23  &  18.5M \\
\midrule
\multirow{2}{*}{($+$) Alignment} & \textbf{Single-Multi} & 11.10 & 5.38 & 1.31 & 8.53  & 0.64 & 0.22 & 0.067 & 2.91 & 12.2M \\
                                 & Multi-Multi     & 11.30 & 5.77 & 1.39 & 8.66  & 0.74 & 0.25 & 0.082 & 3.28 & 12.2M\\
\midrule
\multirow{2}{*}{($+$) Motion hidden state} & Separated & 20.36 & 12.08 & 2.82 & 20.38  & 3.84 & 2.25 & 0.511 & 6.07 &  12.2M   \\          
                                    & \textbf{Shared} & 11.09 & 5.33 & 1.24 & 8.55 & 0.61 & 0.22 & 0.062 & 2.81 & 12.2M \\   
\midrule
\midrule
\multirow{3}{*}{($+$) Inference iterations} & 12  & 11.30 & 5.43 & 1.26 & 8.41  & 0.61 & 0.22 & 0.061 & 2.88 & 12.2M\\
                                & \textbf{20}  & 11.09 & 5.33 & 1.24 & 8.55 & 0.61 & 0.22 & 0.062 & 2.81 & 12.2M\\
                                & 32             & 11.10 & 5.30 & 1.23 & 8.61 & 0.62 & 0.22 & 0.063 & 2.90 & 12.2M\\
\midrule
\multirow{2}{*}{($+$) Inference frames}  & 10        & 11.38 & 5.47 & 1.32 & 8.74 & 0.66 & 0.24 & 0.081 & 2.81 & 12.2M\\
                                   & \textbf{20} & 11.09 & 5.33 & 1.24 & 8.55 & 0.61 & 0.22 & 0.062 & 2.81 & 12.2M\\

\bottomrule
\end{tabular}
\label{tab:ablation}
\end{table}

\subsection{Ablation Studies} \label{Ablation Studies}
    As shown in Tab.~\ref{tab:ablation}, we study the specific components of our approach in isolation and bold the settings used in the final model. The baseline is a per-frame model with 2D $1 \times 5$ convolutions in the updater and shared weights update module. We train the models on SceneFlow \cite{mayer2016large} and Dynamic Replica \cite{karaev2023dynamicstereo} with the hyper-parameters described in Sec.~\ref{Implementation Details}. We evaluate these models on the final pass of Sintel \cite{sintel} and on the test split of Dynamic Replica \cite{karaev2023dynamicstereo}.

\noindent \textbf{Updater convolution.} We first compare 2D and 3D convolutions in the iterative updater (Sec.~\ref{Motion-propagation based Recurrent Unit}). As can be seen from the table, compared with the baseline model, using 3D convolutions benefits the temporal consistency, with a large improvement in TEPE from 1.74 to 1.34. This illustrates the effectiveness of processing  multiple frames simultaneously. We also find that the accuracy of the method with 3D convolutions  decreases slightly. This may be due to the fusion of unaligned features across time. While prior works use $1\times1\times5$ convolutions in the updater, we find it beneficial to enlarge the kernel size of the updater across horizontal dimensions. This results in a general improvement in accuracy and temporal consistency.

\noindent \textbf{Update module.} We conducted experiments on shared and separated weights for the update modules in three resolutions. According to the results for separated weights there is no obvious improvement but the number of parameters is three times larger than for the shared weights. 

\noindent \textbf{Bidirectional alignment.} After adding the bidirectional alignment to the correlation layer, we observed significant performance improvements across all metrics. We experimented with constructing triple-frame cost volumes for both, adjacent aligned left frames and their corresponding aligned right frames (multi-multi). However, this strategy resulted in a performance drop, likely due to the extra noise from more than one reference image. For the task of pixel-level matching, the fixed reference image and the expanded temporal receptive field of the source images led to better results.

\noindent \textbf{Motion hidden state.} In the proposed method, motion propagation is developed to leverage the global temporal information from the whole sequences. The motion hidden state from the previous update module was upscaled to the next one (shared), which proved to be effective, as shown in Tab.~\ref{tab:ablation}. We also report results for separating and re-initializing the motion hidden state in each resolution (separated). However, this leads to an unstable training and a significant performance drop. 

\noindent \textbf{Inference settings.} Different iteration times and number of frames are also investigated. As shown in Tab.~\ref{tab:ablation}, $20$ iterations were optimal for inference. Processing more frames at once during inference improved the results.

\subsection{Parameter, Memory, and MAC Counts}

\begin{figure}[ht]
  \centering
  \includegraphics[width=.99\textwidth]{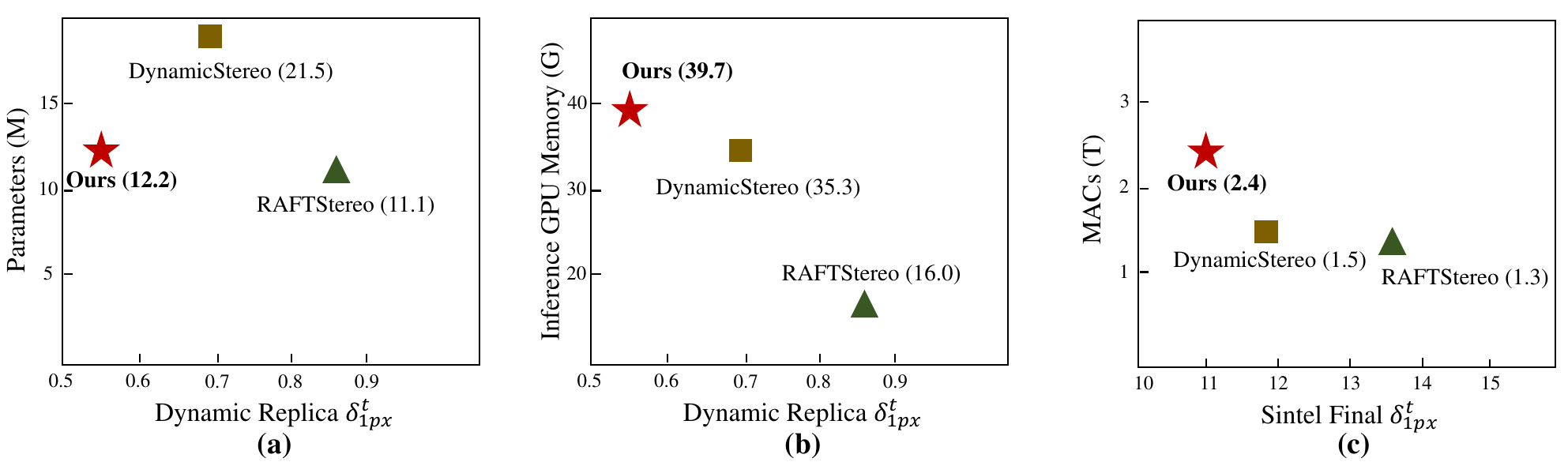}
  \caption{(a) $\delta_{1px}^{t}$ on DR vs. parameters. (b) $\delta_{1px}^{t}$ on DR vs. inference GPU memory (20 frames $\times$ 768 $\times$ 1024). (c) $\delta_{1px}^{t}$ on Sintel vs. MACs (436 $\times$ 1024).}
  \label{fig:efficiency}
\end{figure}

As shown in Fig.~\ref{fig:efficiency}, we compare the methods in terms  of parameters, inference GPU memory and multiply–accumulate (MAC). Our method  strikes a balance in these three criteria while giving the lowest error. The main overheads of our method in inference memory and MAC are from the optical flow module.

\section{Conclusion and Future Work}
In this paper, we show that the bidirectional alignment improves the consistency of dynamic stereo matching. Based on this, we propose BiDAStereo, which integrates a triple-frame correlation layer and a motion-propagation recurrent unit to effectively extract both local and global temporal cues. Experimental results show that our approach performs well on various datasets, especially on out-of-domain scenes. A common limitation among our work and existing works is the inability to proactively distinguish between dynamic and static areas, which is the key to ensuring consistency. Moving forward, our focus lies in exploring how to integrate our method with explicit geometry priors to enhance the performance and in developing a lightweight version of the model.

%
%
\bibliographystyle{splncs04}
\bibliography{main}

\begin{thebibliography}{10}
\providecommand{\url}[1]{\texttt{#1}}
\providecommand{\urlprefix}{URL }
\providecommand{\doi}[1]{https://doi.org/#1}

\bibitem{azuma1997survey}
Azuma, R.T.: A survey of augmented reality. Presence: teleoperators \& virtual environments  \textbf{6}(4),  355--385 (1997)

\bibitem{bao2020instereo2k}
Bao, W., Wang, W., Xu, Y., Guo, Y., Hong, S., Zhang, X.: Instereo2k: a large real dataset for stereo matching in indoor scenes. Science China Information Sciences  \textbf{63}(11),  1--11 (2020)

\bibitem{birchfield1999depth}
Birchfield, S., Tomasi, C.: Depth discontinuities by pixel-to-pixel stereo. IJCV  \textbf{35}(3),  269--293 (1999)

\bibitem{bleyer2011patchmatch}
Bleyer, M., Rhemann, C., Rother, C.: Patchmatch stereo-stereo matching with slanted support windows. In: Bmvc. vol.~11, pp. 1--11 (2011)

\bibitem{boykov2001fast}
Boykov, Y., Veksler, O., Zabih, R.: Fast approximate energy minimization via graph cuts. IEEE TPAMI  \textbf{23}(11),  1222--1239 (2001)

\bibitem{sintel}
Butler, D.J., Wulff, J., Stanley, G.B., Black, M.J.: A naturalistic open source movie for optical flow evaluation. In: ECCV. pp. 611--625 (2012)

\bibitem{chang2018pyramid}
Chang, J.R., Chen, Y.S.: Pyramid stereo matching network. In: CVPR. pp. 5410--5418 (2018)

\bibitem{chang2023domain}
Chang, T., Yang, X., Zhang, T., Wang, M.: Domain generalized stereo matching via hierarchical visual transformation. In: Proceedings of the IEEE/CVF Conference on Computer Vision and Pattern Recognition. pp. 9559--9568 (2023)

\bibitem{cheng2024stereo}
Cheng, Z., Yang, J., Li, H.: Stereo matching in time: 100+ fps video stereo matching for extended reality. In: Proceedings of the IEEE/CVF Winter Conference on Applications of Computer Vision. pp. 8719--8728 (2024)

\bibitem{deschaud2021kitti}
Deschaud, J.E.: Kitti-carla: a kitti-like dataset generated by carla simulator. arXiv preprint arXiv:2109.00892  (2021)

\bibitem{desouza2002vision}
DeSouza, G.N., Kak, A.C.: Vision for mobile robot navigation: A survey. IEEE transactions on pattern analysis and machine intelligence  \textbf{24}(2),  237--267 (2002)

\bibitem{kitti}
Geiger, A., Lenz, P., Urtasun, R.: Are we ready for autonomous driving? the kitti vision benchmark suite. In: CVPR. pp. 3354--3361 (2012)

\bibitem{geiger2011stereoscan}
Geiger, A., Ziegler, J., Stiller, C.: Stereoscan: Dense 3d reconstruction in real-time. In: 2011 IEEE intelligent vehicles symposium (IV). pp. 963--968. Ieee (2011)

\bibitem{guo2019group}
Guo, X., Yang, K., Yang, W., Wang, X., Li, H.: Group-wise correlation stereo network. In: CVPR. pp. 3273--3282 (2019)

\bibitem{hartley2003multiple}
Hartley, R., Zisserman, A.: Multiple view geometry in computer vision. Cambridge university press (2003)

\bibitem{hirschmuller2002real}
Hirschm{\"u}ller, H., Innocent, P.R., Garibaldi, J.: Real-time correlation-based stereo vision with reduced border errors. IJCV  \textbf{47}(1),  229--246 (2002)

\bibitem{Jing_2023_ICCV}
Jing, J., Li, J., Xiong, P., Liu, J., Liu, S., Guo, Y., Deng, X., Xu, M., Jiang, L., Sigal, L.: Uncertainty guided adaptive warping for robust and efficient stereo matching. In: Proceedings of the IEEE/CVF International Conference on Computer Vision (ICCV). pp. 3318--3327 (October 2023)

\bibitem{karaev2023dynamicstereo}
Karaev, N., Rocco, I., Graham, B., Neverova, N., Vedaldi, A., Rupprecht, C.: Dynamicstereo: Consistent dynamic depth from stereo videos. In: Proceedings of the IEEE/CVF Conference on Computer Vision and Pattern Recognition. pp. 13229--13239 (2023)

\bibitem{kendall2017end}
Kendall, A., Martirosyan, H., Dasgupta, S., Henry, P., Kennedy, R., Bachrach, A., Bry, A.: End-to-end learning of geometry and context for deep stereo regression. In: CVPR. pp. 66--75 (2017)

\bibitem{adam}
Kingma, D.P., Ba, J.: Adam: A method for stochastic optimization. arXiv preprint arXiv:1412.6980  (2014)

\bibitem{klaus2006segment}
Klaus, A., Sormann, M., Karner, K.: Segment-based stereo matching using belief propagation and a self-adapting dissimilarity measure. In: ICPR. vol.~3, pp. 15--18 (2006)

\bibitem{li2022practical}
Li, J., Wang, P., Xiong, P., Cai, T., Yan, Z., Yang, L., Liu, J., Fan, H., Liu, S.: Practical stereo matching via cascaded recurrent network with adaptive correlation. In: Proceedings of the IEEE/CVF Conference on Computer Vision and Pattern Recognition. pp. 16263--16272 (2022)

\bibitem{li2023temporally}
Li, Z., Ye, W., Wang, D., Creighton, F.X., Taylor, R.H., Venkatesh, G., Unberath, M.: Temporally consistent online depth estimation in dynamic scenes. In: Proceedings of the IEEE/CVF winter conference on applications of computer vision. pp. 3018--3027 (2023)

\bibitem{lipson2021raft}
Lipson, L., Teed, Z., Deng, J.: Raft-stereo: Multilevel recurrent field transforms for stereo matching. arXiv preprint arXiv:2109.07547  (2021)

\bibitem{liu2022video}
Liu, H., Ruan, Z., Zhao, P., Dong, C., Shang, F., Liu, Y., Yang, L., Timofte, R.: Video super-resolution based on deep learning: a comprehensive survey. Artificial Intelligence Review  \textbf{55}(8),  5981--6035 (2022)

\bibitem{mayer2016large}
Mayer, N., Ilg, E., Hausser, P., Fischer, P., Cremers, D., Dosovitskiy, A., Brox, T.: A large dataset to train convolutional networks for disparity, optical flow, and scene flow estimation. In: CVPR. pp. 4040--4048 (2016)

\bibitem{menze2015object}
Menze, M., Geiger, A.: Object scene flow for autonomous vehicles. In: CVPR. pp. 3061--3070 (2015)

\bibitem{pang2017cascade}
Pang, J., Sun, W., Ren, J.S., Yang, C., Yan, Q.: Cascade residual learning: A two-stage convolutional neural network for stereo matching. In: CVPRW. pp. 887--895 (2017)

\bibitem{pang2018zoom}
Pang, J., Sun, W., Yang, C., Ren, J., Xiao, R., Zeng, J., Lin, L.: Zoom and learn: Generalizing deep stereo matching to novel domains. In: CVPR. pp. 2070--2079 (2018)

\bibitem{rao2023masked}
Rao, Z., Xiong, B., He, M., Dai, Y., He, R., Shen, Z., Li, X.: Masked representation learning for domain generalized stereo matching. In: Proceedings of the IEEE/CVF Conference on Computer Vision and Pattern Recognition. pp. 5435--5444 (2023)

\bibitem{scharstein2014high}
Scharstein, D., Hirschm{\"u}ller, H., Kitajima, Y., Krathwohl, G., Ne{\v{s}}i{\'c}, N., Wang, X., Westling, P.: High-resolution stereo datasets with subpixel-accurate ground truth. In: German Conference on Pattern Recognition. pp. 31--42 (2014)

\bibitem{scharstein2002taxonomy}
Scharstein, D., Szeliski, R.: A taxonomy and evaluation of dense two-frame stereo correspondence algorithms. IJCV  \textbf{47}(1),  7--42 (2002)

\bibitem{middlebury}
Scharstein, D., Szeliski, R.: A taxonomy and evaluation of dense two-frame stereo correspondence algorithms. IJCV  \textbf{47}(1),  7--42 (2002)

\bibitem{eth3d}
Schops, T., Schonberger, J.L., Galliani, S., Sattler, T., Schindler, K., Pollefeys, M., Geiger, A.: A multi-view stereo benchmark with high-resolution images and multi-camera videos. In: CVPR. pp. 3260--3269 (2017)

\bibitem{airsim2017fsr}
Shah, S., Dey, D., Lovett, C., Kapoor, A.: Airsim: High-fidelity visual and physical simulation for autonomous vehicles. In: Field and Service Robotics (2017), \url{https://arxiv.org/abs/1705.05065}

\bibitem{shen2021cfnet}
Shen, Z., Dai, Y., Rao, Z.: Cfnet: Cascade and fused cost volume for robust stereo matching. In: CVPR. pp. 13906--13915 (2021)

\bibitem{smith2019super}
Smith, L.N., Topin, N.: Super-convergence: Very fast training of neural networks using large learning rates. In: Artificial intelligence and machine learning for multi-domain operations applications. vol. 11006, pp. 369--386. SPIE (2019)

\bibitem{song2021adastereo}
Song, X., Yang, G., Zhu, X., Zhou, H., Wang, Z., Shi, J.: Adastereo: a simple and efficient approach for adaptive stereo matching. In: CVPR. pp. 10328--10337 (2021)

\bibitem{sun2003stereo}
Sun, J., Zheng, N.N., Shum, H.Y.: Stereo matching using belief propagation. IEEE TPAMI  \textbf{25}(7),  787--800 (2003)

\bibitem{tankovich2021hitnet}
Tankovich, V., Hane, C., Zhang, Y., Kowdle, A., Fanello, S., Bouaziz, S.: Hitnet: Hierarchical iterative tile refinement network for real-time stereo matching. In: CVPR. pp. 14362--14372 (2021)

\bibitem{teed2020raft}
Teed, Z., Deng, J.: Raft: Recurrent all-pairs field transforms for optical flow. In: ECCV. pp. 402--419 (2020)

\bibitem{teed2021raft}
Teed, Z., Deng, J.: Raft-3d: Scene flow using rigid-motion embeddings. In: Proceedings of the IEEE/CVF conference on computer vision and pattern recognition. pp. 8375--8384 (2021)

\bibitem{fallingthings}
Tremblay, J., To, T., Birchfield, S.: Falling things: A synthetic dataset for 3d object detection and pose estimation. In: CVPRW. pp. 2038--2041 (2018)

\bibitem{van2002hierarchical}
Van~Meerbergen, G., Vergauwen, M., Pollefeys, M., Van~Gool, L.: A hierarchical symmetric stereo algorithm using dynamic programming. IJCV  \textbf{47}(1),  275--285 (2002)

\bibitem{xu2023iterative}
Xu, G., Wang, X., Ding, X., Yang, X.: Iterative geometry encoding volume for stereo matching. In: Proceedings of the IEEE/CVF Conference on Computer Vision and Pattern Recognition. pp. 21919--21928 (2023)

\bibitem{xu2022acvnet}
Xu, G., Wang, Y., Cheng, J., Tang, J., Yang, X.: Accurate and efficient stereo matching via attention concatenation volume. arXiv preprint arXiv:2209.12699  (2022)

\bibitem{xu2020aanet}
Xu, H., Zhang, J.: Aanet: Adaptive aggregation network for efficient stereo matching. In: CVPR. pp. 1959--1968 (2020)

\bibitem{yang2019hierarchical}
Yang, G., Manela, J., Happold, M., Ramanan, D.: Hierarchical deep stereo matching on high-resolution images. In: CVPR. pp. 5515--5524 (2019)

\bibitem{yang2008stereo}
Yang, Q., Wang, L., Yang, R., Stew{\'e}nius, H., Nist{\'e}r, D.: Stereo matching with color-weighted correlation, hierarchical belief propagation, and occlusion handling. IEEE TPAMI  \textbf{31}(3),  492--504 (2008)

\bibitem{yao2018mvsnet}
Yao, Y., Luo, Z., Li, S., Fang, T., Quan, L.: Mvsnet: Depth inference for unstructured multi-view stereo. In: Proceedings of the European conference on computer vision (ECCV). pp. 767--783 (2018)

\bibitem{zbontar2015computing}
Zbontar, J., LeCun, Y.: Computing the stereo matching cost with a convolutional neural network. In: CVPR. pp. 1592--1599 (2015)

\bibitem{zhang2019ga}
Zhang, F., Prisacariu, V., Yang, R., Torr, P.H.: Ga-net: Guided aggregation net for end-to-end stereo matching. In: CVPR. pp. 185--194 (2019)

\bibitem{zhang2023temporalstereo}
Zhang, Y., Poggi, M., Mattoccia, S.: Temporalstereo: Efficient spatial-temporal stereo matching network. In: 2023 IEEE/RSJ International Conference on Intelligent Robots and Systems (IROS). pp. 9528--9535. IEEE (2023)

\bibitem{zhong2018open}
Zhong, Y., Li, H., Dai, Y.: Open-world stereo video matching with deep rnn. In: Proceedings of the European Conference on Computer Vision (ECCV). pp. 101--116 (2018)

\end{thebibliography}
\end{document}